\documentclass[sigconf, balance=true]{acmart}
\AtBeginDocument{%
  \providecommand\BibTeX{{%
    \normalfont B\kern-0.5em{\scshape i\kern-0.25em b}\kern-0.8em\TeX}}}

\setcopyright{acmcopyright}
\copyrightyear{2018}
\acmYear{2018}
\acmDOI{10.1145/1122445.1122456}

\acmConference[Woodstock '18]{Woodstock '18: ACM Symposium on Neural
  Gaze Detection}{June 03--05, 2018}{Woodstock, NY}
\acmBooktitle{Woodstock '18: ACM Symposium on Neural Gaze Detection,
  June 03--05, 2018, Woodstock, NY}
\acmPrice{15.00}
\acmISBN{978-1-4503-XXXX-X/18/06}

\usepackage{caption}
\usepackage{enumitem}
\usepackage{tabu}
\usepackage{rotating}
\usepackage{amsmath}
\usepackage{tikz}
\usepackage{capt-of,etoolbox}
\usepackage{microtype}
\usepackage{bm}
\usepackage{soul}
\usepackage{wrapfig}
\usepackage{graphbox}
\usepackage{tcolorbox}
\usepackage{cleveref}
\usepackage{mathtools}
\usepackage{balance}
\usepackage[bb=boondox]{mathalfa}
\usepackage[utf8x]{inputenc}

\definecolor{orange}{RGB}{250,130,49}
\definecolor{red}{RGB}{234,59,90}
\definecolor{agreen}{RGB}{74, 198, 148}
\definecolor{purple}{RGB}{158, 62, 177}
\definecolor{darkpurple}{RGB}{170, 70, 210}
\definecolor{aqua}{RGB}{87, 180, 181}
\definecolor{lightblue}{RGB}{72, 123, 232}
\definecolor{hotpink}{RGB}{255, 83, 115}
\definecolor{teal}{RGB}{90, 200, 250}
\definecolor{linkColor}{RGB}{0, 128, 229}
\definecolor{lightgreen}{RGB}{33, 222, 128}
\definecolor{gray}{RGB}{75, 101, 132}

\definecolor{germanred}{RGB}{234, 59, 90}
\definecolor{germanorange}{RGB}{250, 130, 49}
\definecolor{germanyellow}{RGB}{254, 211, 48}
\definecolor{germangreen}{RGB}{32, 191, 107}
\definecolor{germanblue}{RGB}{0, 128, 229}
\definecolor{germanviolet}{RGB}{56, 103, 214}
\definecolor{germanpurple}{RGB}{136, 84, 208}

\newcommand{\link}[1]{{\color{linkColor}\href{#1}{\textbf{\texttt{#1}}}}}

\crefname{figure}{fig.}{fig.}
\Crefname{figure}{Fig.}{Fig.}
\crefname{section}{\S}{\S}

\newcommand{\tool}{\textsc{\textsf{GAM Changer}}}

\newtoggle{inheader}
\newcommand{\canvasview}{\iftoggle{inheader}{GAM Canvas}{\textit{GAM Canvas}}}
\newcommand{\metricview}{\iftoggle{inheader}{Metric Panel}{\textit{Metric Panel}}}
\newcommand{\featureview}{\iftoggle{inheader}{Feature Panel}{\textit{Feature Panel}}}
\newcommand{\historyview}{\iftoggle{inheader}{History Panel}{\textit{History Panel}}}
\newcommand{\toolbar}{\iftoggle{inheader}{Context Toolbar}{\textit{Context Toolbar}}}
\newcommand{\statusbar}{\iftoggle{inheader}{Status Bar}{\textit{Status Bar}}}
\newcommand{\scopeswitch}{\iftoggle{inheader}{Scope Switch}{\textit{Scope Switch}}}

\DeclarePairedDelimiter\set\{\}

\definecolor{soulblue}{RGB}{194, 220, 242}
\definecolor{soulorange}{RGB}{255, 223, 179}
\colorlet{soulgermanblue}{germanblue!30}
\newcommand{\bluehl}[1]{{\sethlcolor{soulblue}\hl{#1}}}
\newcommand{\orangehl}[1]{{\sethlcolor{soulorange}\hl{#1}}}

\newcommand{\inlinefig}[2][10]{\protect\includegraphics[align=c, height=#1pt]{./figures/#2.pdf}}

\definecolor{tagbordercolor}{rgb}{0.8, 0.8, 0.8}
\definecolor{tagbgcolor}{rgb}{0.9, 0.9, 0.9}

\newtcbox{\tagg}{nobeforeafter, colframe=tagbordercolor,
colback=tagbgcolor, boxrule=0.5pt, arc=1pt,
  boxsep=0pt,left=2pt,right=2pt,top=1.5pt,bottom=2pt,tcbox raise base}

\setcopyright{none}
\renewcommand\footnotetextcopyrightpermission[1]{}

\begin{document}

\title{\tool{}: Editing Generalized Additive Models with Interactive Visualization}

\newcommand{\authorgap}{\hspace{3pt}}

\author{Zijie J. Wang$^{1}$ \authorgap Alex Kale$^{2}$ \authorgap Harsha Nori$^{3}$ \authorgap Peter Stella$^{4}$ \authorgap Mark Nunnally$^{4}$ \authorgap Duen Horng Chau$^{1}$ \authorgap Mihaela Vorvoreanu$^{3}$ \authorgap Jennifer Wortman Vaughan$^{3}$ \authorgap Rich Caruana$^{3}$
}
\affiliation{
  \institution{$^{1}$Georgia Tech \authorgap $^{2}$University of Washington \authorgap $^{3}$Microsoft Research \authorgap $^{4}$NYU Langone Health}
  \country{}
}

\renewcommand{\shortauthors}{Wang, et al.}

\begin{abstract}
  Recent strides in interpretable machine learning (ML) research reveal that models exploit undesirable patterns in the data to make predictions, which potentially causes harms in deployment.
  However, it is unclear how we can fix these models.
  We present our ongoing work, \tool{}, an open-source interactive system to help data scientists and domain experts easily and responsibly edit their Generalized Additive Models (GAMs).
  With novel visualization techniques, our tool puts interpretability into action---empowering human users to analyze, validate, and align model behaviors with their knowledge and values.
  Built using modern web technologies, our tool runs locally in users' computational notebooks or web browsers without requiring extra compute resources, lowering the barrier to creating more responsible ML models.
  \tool{} is available at \link{https://interpret.ml/gam-changer}.
\end{abstract}

\begin{teaserfigure}
  \includegraphics[width=\textwidth]{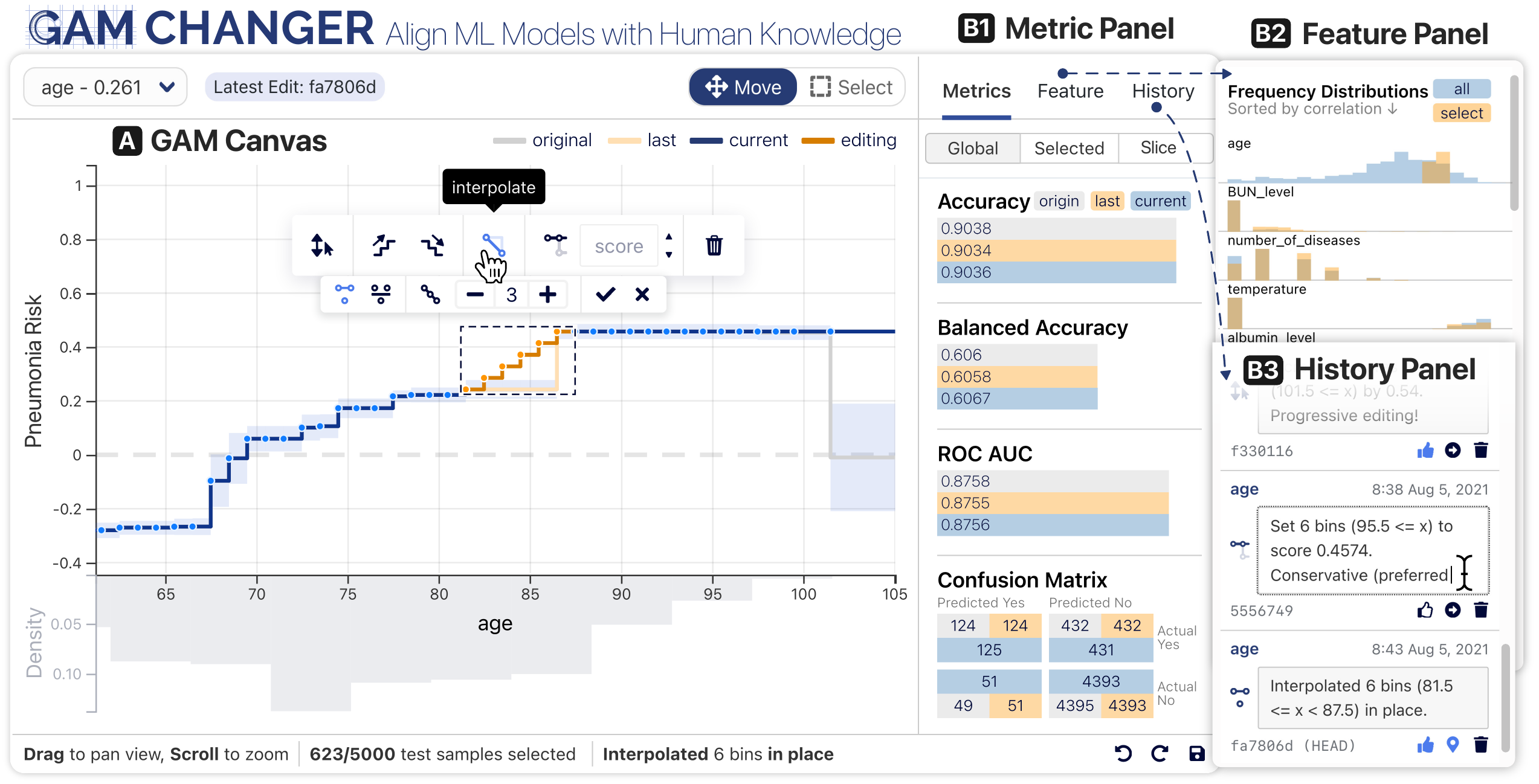}
  \Description{User interface for \tool{}, featuring four tightly integrated views:
    \canvasview{}, \metricview{}, \featureview{}, and \historyview{}
  }
  \caption{
    \tool{} empowers domain experts and data scientists to easily and responsibly
    align model behaviors with their domain knowledge and values, via direct manipulation of GAM model weights.
    For example, \inlinefig[10]{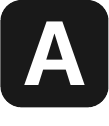} the \canvasview{} enables doctors to interpolate the predicted risk of dying from pneumonia to match their domain knowledge of a gradual risk increase from age \texttt{81} to age \texttt{87}.
    \tool{} promotes accountable editing and elucidates potential tradeoffs induced by the edits.
    \inlinefig[10]{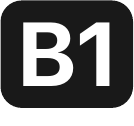} The \metricview{} provides real time feedback on model performance.
    \inlinefig[10]{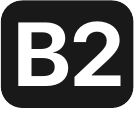} The \featureview{} helps users identify characteristics of affected samples and promotes awareness of fairness issues.
    To enable reversible transparent model edits, \inlinefig[10]{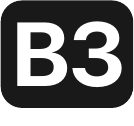} the \historyview{} allows the doctor to compare and revert changes, as well as document their motivations and editing contexts.
  }
  \label{fig:teaser}
\end{teaserfigure}

\maketitle
\pagestyle{plain}

\section{Introduction}

\begin{figure}[tb]
  \includegraphics[width=\linewidth]{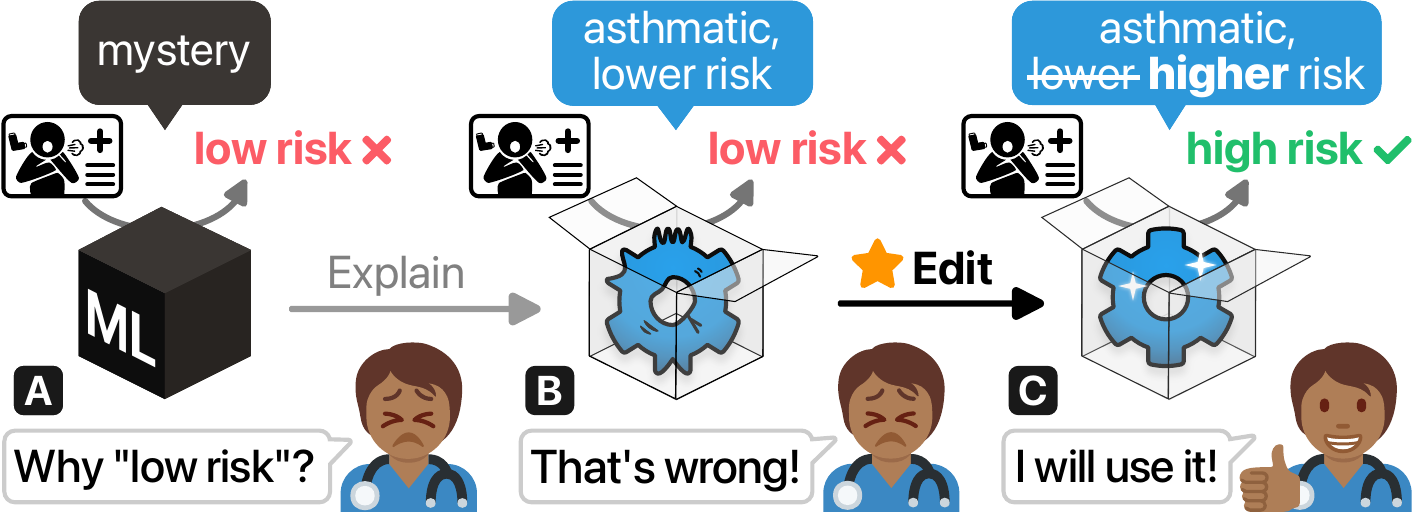}
  \Description{User interface for \tool{}, featuring four tightly integrated views:
    \canvasview{}, \metricview{}, \featureview{}, and \historyview{}
  }
  \caption{
    \inlinefig[10]{a2} Doctors often hesitate to trust ML models as they cannot interpret how the models make predictions.
    \inlinefig[10]{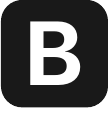} With interpretable ML, researchers and doctors discover models can learn unexpected patterns, potentially causing harm in deployment.
    \inlinefig[10]{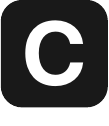} Model editing turns interpretability into actions, enabling doctors to repair ML models and align model behaviors with their domain knowledge.
  }
  \label{fig:motivation}
\end{figure}

It is crucial to understand how machine learning (ML) models used in high-stakes settings (e.g., healthcare, finance, and criminal justice) make predictions~(\autoref{fig:motivation}A).
Recently, researchers have made substantial efforts to make ML models interpretable~\cite[e.g.,][]{ribeiroWhyShouldTrust2016,lundbergUnifiedApproachInterpreting2017,caruanaIntelligibleModelsHealthCare2015}.
Yet, not as much research focuses on what do we do with these model explanations.
In practice, data scientists and domain experts often \textit{compare} model explanations with their knowledge~\cite{hongHumanFactorsModel2020}.
If the model uses expected patterns to make predictions, people will feel more confident to deploy it to solve real problems.
Sometimes, ML interpretability can uncover hidden relationships in the data---helping people gain insights about the problems they want to tackle.

Other times, however, ML interpretability reveals models' exploitation of dangerous patterns in the data to make predictions.
These patterns might be an accurate reflection of real-world phenomena, but leaving them untouched can cause serious harm in deployment.
For example, ML models that decide mortgage approvals exhibit biases against loan applicants of color~\cite{martinezSecretBiasHidden2021}.
These biases reflect and are inherited from the real world~\cite{bartlettConsumerLendingDiscriminationFinTech2019}.
In healthcare, state-of-the-art models show that having asthma lowers the risk of dying from pneumonia~(\autoref{fig:motivation}B), where researchers suspect it is because asthmatic patients receive care earlier~\cite{caruanaIntelligibleModelsHealthCare2015}.
If we use this flawed model to make hospital admission decisions, asthmatic patients are likely to miss the care they need.
Interpretability helps us identify these dangerous patterns, but how can we take a step further and use model explanations to \textit{improve}~(\autoref{fig:motivation}C) ML models?

To answer this question, we are developing \textbf{\tool{}}~(\autoref{fig:teaser}): an interactive visualization system that empowers data scientists and domain experts to easily and responsibly edit the weights of \textit{generalized additive models} (GAMs), the state-of-the-art interpretable ML model for tabular data~\cite{hastieGeneralizedAdditiveModels1999, caruanaIntelligibleModelsHealthCare2015}.
We iteratively design this tool by continuously integrating feedback from ML and human-computer interaction (HCI) researchers, data scientists, and doctors.
In this ongoing work, we contribute:

\begin{itemize}[topsep=1mm, itemsep=0mm, parsep=1mm, leftmargin=3mm]

\item \textbf{\tool{}, a novel interactive system} that empowers data scientists and domain experts to edit GAMs to align model behaviors with their knowledge and values.
Advancing over prior visualization work for interpretable ML~\cite{hohmanGamutDesignProbe2019,wangDodrioExploringTransformer2021}, our tool is the first system that enables users to directly modify ML models with model explanations through interactive visualization.

\item \textbf{Responsible and novel visualization design} through a participatory and iterative process with doctors and data scientists.
Inspired by popular image editing and illustration tools, \tool{} adapts familiar \textit{Direct Manipulation}~\cite{shneidermanDirectManipulationStep1983} user interfaces to edit complex ML models.
Accessible model editing empowers users to exercise their human agency but demands caution, as modifications of high-stake models have serious consequences.
Therefore, guarding against harmful edits is our top design priority:
we leverage \textit{continuous feedback} and \textit{reversible actions} to elucidate editing effects and promote accountable edits.

\item \textbf{An open-source\footnote{\tool{} code: \link{https://github.com/interpretml/gam-changer}}, web-based implementation} that broadens people's access to creating more accountable ML models and exercising their human agency in a world penetrated by AI systems.
We develop \tool{} with modern web technologies such as WebAssembly\footnote{WebAssembly: \link{https://webassembly.org}}.
Therefore, anyone can access our tool directly in their web browser or computational notebooks and edit ML models with real datasets at scale.
For a demo video of \tool{}, visit \link{https://youtu.be/2gVSoPoSeJ8}.

\end{itemize}

We hope our work helps emphasize the importance of human agency in responsible ML research, inspiring future work in human-AI interaction and actionable ML interpretability.

\section{Background and Related Work}

\newcommand{\mcolor}[2]{\textcolor{#1}{#2}}
\newcommand{\tcolor}[2]{\textcolor{#1}{#2}}

Generalized additive models (GAMs) have emerged as one of the most popular model classes among today's data science community.
GAMs' predictive performance is on par with more complex, state-of-the-art models, yet GAMs remain simple enough for humans to understand its decision process~\cite{changHowInterpretableTrustworthy2021}.
Given an \tcolor{germanorange}{input} $\mcolor{germanorange}{x \in \mathbb{R}^{M}}$ with \tcolor{germanorange}{$M$ features} and a \tcolor{germangreen}{target} $\mcolor{germangreen}{y \in \mathbb{R}}$, a GAM with a \tcolor{germanpurple}{link function} $\mcolor{germanpurple}{g}$ and \tcolor{germanblue}{shape function} $\mcolor{germanblue}{f_j}$ for each feature $j \in \set{1, 2, \dots, M}$ can be written as:
\begin{equation}
    \label{equation:gam}
    \mcolor{germanpurple}{g \left(\mcolor{germangreen}{y}\right)} = \mcolor{gray}{\beta_0} + \mcolor{germanblue}{f_1 \left(\mcolor{germanorange}{x_1}\right)} + \mcolor{germanblue}{f_2 \left(\mcolor{germanorange}{x_2}\right)} + \cdots + \mcolor{germanblue}{f_M \left(\mcolor{germanorange}{x_M}\right)}
\end{equation}

The \tcolor{germanpurple}{link function} is determined by the task.  For example, in binary classification, $\mcolor{germanpurple}{g}$ is \tcolor{germanpurple}{logit}.
In \autoref{equation:gam}, $\mcolor{gray}{\beta_0}$ represents \tcolor{gray}{the intercept constant}.
There are many options for the \tcolor{germanblue}{shape functions} $\mcolor{germanblue}{f_j}$, such as \tcolor{germanblue}{splines~\cite{hastieGeneralizedAdditiveModels1999}}, \tcolor{germanblue}{gradient-boosted trees~\cite{caruanaIntelligibleModelsHealthCare2015}}, and \tcolor{germanblue}{neural networks~\cite{agarwalNeuralAdditiveModels2020}}.
Some GAMs also support pair-wise interaction terms $\mcolor{germanblue}{f_{ij}(\mcolor{germanorange}{x_i, x_j})}$.  Different GAM variants come with different training methods, but once trained, they all have the same form.
The interpretability and editability of GAMs stem from the fact that people can visualize and modify each \tcolor{germanorange}{feature $x_j$}'s contribution score to the model's prediction by inspecting and adjusting the \tcolor{germanblue}{shape function} $\mcolor{germanblue}{f_j}$.
The contribution score is measured by the output of \tcolor{germanblue}{$f_j(\mcolor{germanorange}{x_j})$}.
Since GAMs are additive, we can edit different \tcolor{germanblue}{shape functions} independently.

Besides \textit{glass-box} models like GAMs that are inherently interpretable~\cite[e.g.,][]{zengInterpretableClassificationModels2017,lakkarajuInterpretableDecisionSets2016}, ML researchers have developed post hoc explanation methods to interpret any \textit{black-box} models~\cite[e.g.,][]{ribeiroWhyShouldTrust2016,lundbergUnifiedApproachInterpreting2017}.
HCI researchers study how to communicate model explanations~\cite{kaurInterpretingInterpretabilityUnderstanding2020}, and develop visual analytics systems to help users interactively analyze ML models~\cite{hohmanGamutDesignProbe2019, wangDodrioExploringTransformer2021,wexlerWhatIfToolInteractive2019}.
Our work advances the interpretable ML landscape in \textbf{making interpretability actionable}.
Although research shows that modifying model outputs can lead to greater trust and better human-AI teaming performance~\cite{dietvorstOvercomingAlgorithmAversion2018}, there has been little work on how we can leverage interpretability to adjust models.
Most existing work~\cite{bauUnderstandingRoleIndividual2020,bauIdentifyingControllingImportant2019,suauFindingExpertsTransformer2020} relies on black-box models and post hoc explanations---users can only affect a small subset of model behaviors and modifications are likely to have unknown effects.
Grounded on accurate and complete interpretations provided by glass-box models, \tool{} is the \textbf{first framework} that enables users to have total control of their model behavior and recognize the editing effects,
so that they can easily and safely improve ML models in solving high-stake problems.

\begin{figure*}[tb]
  \includegraphics[width=\linewidth]{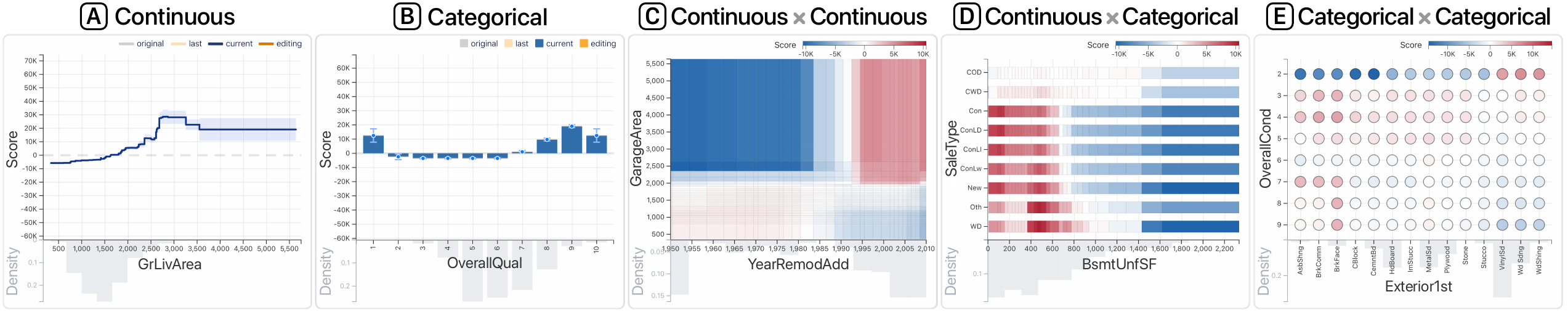}
  \Description{User interface for \tool{}, featuring four tightly integrated views:
    \canvasview{}, \metricview{}, \featureview{}, and \historyview{}
  }
  \caption{
    The \canvasview{} illustrates the model decision process by plotting the trained shape functions across features.
    As a GAM's inference varies by feature type, we apply different visualization designs for different feature types.
    \inlinefig[10]{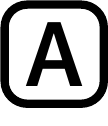} Line charts visualize the contribution of continuous variables.
    \inlinefig[10]{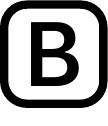} Bar charts show the score of each level in categorical variables.
    \inlinefig[10]{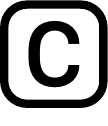} \& \inlinefig[10]{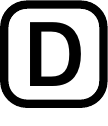} Heatmaps describe interaction effects involving continuous variables.
    \inlinefig[10]{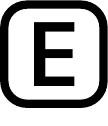} Scatter plots use a grid to explain the interaction effect of two categorical variables.
    For univariate features, the x-axis encodes the input feature $x_j$ and the y-axis represents the output of the shape function $f_j(x_j)$.
    For pair-wise interactions, the axes encode two features, and we use a diverging color scale to represent the contribution scores.
    These examples are from a GAM trained to predict Iowa real estate prices~\cite{decockAmesIowaAlternative2011}.
  }
  \label{fig:gallery}
\end{figure*}

\section{System Design and Implementation}

To lower the barrier to controlling ML model behaviors, \tool{} employs familiar direct manipulation user interface patterns to edit the parameters of GAMs~(\autoref{sec:canvasview}).
By providing real-time feedback about the model performance~(\autoref{sec:metricview}) and potential effect parity across features~(\autoref{sec:featureview}), our tool encourages data scientists and domain experts to edit models responsibly.
All edits are reversible, and users can document and compare their edits~(\autoref{sec:historyview}).
To start \tool{}, users can use our provided functions to extract GAM parameters from a popular GAM library~\cite{noriInterpretMLUnifiedFramework2019} and test samples from the data (either a validation set or a subset of training data).

\vspace{-10pt}
\toggletrue{inheader}
\subsection{\canvasview{}}
\label{sec:canvasview}
\togglefalse{inheader}

\setlength{\columnsep}{8pt}%
\setlength{\intextsep}{0pt}%
\begin{wrapfigure}{R}{0.14\textwidth}
  \vspace{0pt}
  \centering
  \includegraphics[width=0.14\textwidth]{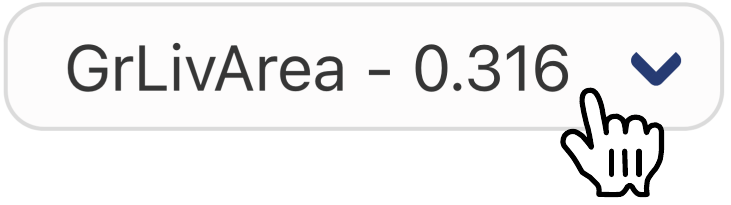}
\end{wrapfigure}

The \canvasview{}~(\autoref{fig:teaser}A) is the primary view of \tool{}, where we visualize one \tcolor{germanorange}{input feature $\mcolor{germanorange}{x_j}$}'s contribution to the model's prediction by plotting its \tcolor{germanblue}{shape function} $\mcolor{germanblue}{f_j(} \mcolor{germanorange}{x_j} \mcolor{germanblue}{)}$.
As GAMs support continuous and categorical features, as well as their two-way interactions, we design separate visualization for each variable type, featuring line chart, bar chart, heatmaps, and scatter plots~(\autoref{fig:gallery}).
Users can use the feature selection drop-down to smoothly transition across features.
When started, the tool shows the feature with the highest importance score, computed by the weighted average of a feature's absolute contribution scores.

\textbf{Shape function visualizations.}
Modern GAMs usually discretize continuous variables into finite bins, so that shape functions can easily learn complex non-linear relationships.
Therefore, the output of shape functions is a continuous piecewise constant function.
We use a dot to show the start of each bin and a line to encode the bin's constant score~(\autoref{fig:teaser}A).
For categorical features, we represent each bin as a bar whose height encodes the bin's score~(\autoref{fig:gallery}B).
To help users keep track of their edits, we color code the lines and bins by the editing sequence: the original function, function in the last edit, and the current function.
We also use histograms to visualize the bin count distribution in the training data.

\textbf{Contribution baseline.}
To help users more easily interpret GAMs, we re-center the contribution scores by adjusting the intercept constant $f_0$, so that the mean prediction for each feature has a zero score across the training data.
Therefore, a positive contribution score suggests that the feature positively affects the prediction, and vice versa.
Consider a GAM trained to predict house prices~(\autoref{fig:gallery}A), if the living area is larger than 2000 square feet, it increases the predicted house price, while areas lower than 2000 decrease the predicted value compared with average.
We highlight the $0$-baseline as a thick dashed line in our visualizations.

\setlength{\columnsep}{6pt}%
\setlength{\intextsep}{0pt}%
\begin{wrapfigure}{R}{0.15\textwidth}
  \vspace{0pt}
  \centering
  \includegraphics[width=0.15\textwidth]{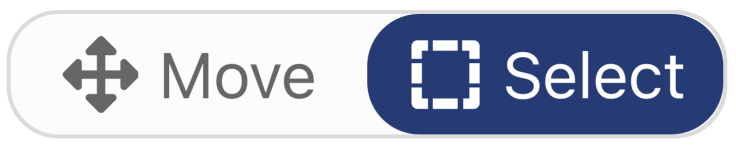}
\end{wrapfigure}

\textbf{Interaction modes.}
In the \canvasview{}, users can switch between \textit{Move Mode} and \textit{Selection Mode} by clicking the mode toggle button.
In the Move Mode, users can use \textit{zoom-and-pan} to control their view portion and focus on analyzing an interesting region in the GAM visualization.
In the Selection Mode, users can use \textit{marquee selection}~(\autoref{fig:teaser}A) to pick a subset of bins (or bars for categorical features) to edit.

\begin{figure}[b]
  \includegraphics[width=0.9\linewidth]{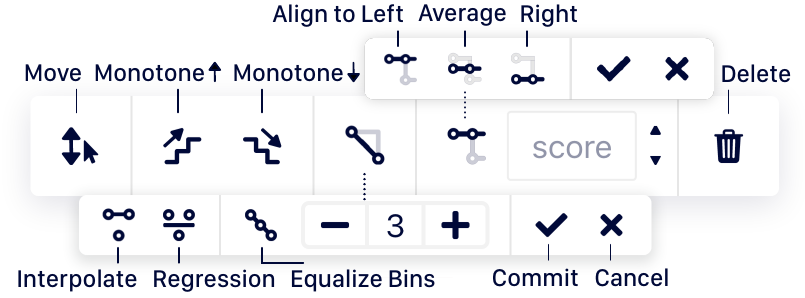}
  \Description{User interface for \tool{}, featuring four tightly integrated views:
    \canvasview{}, \metricview{}, \featureview{}, and \historyview{}
  }
  \caption{
    The \toolbar{} for continuous variable enables users to edit GAMs with a variety of editing methods.
    For example, one user can use the Move tool \inlinefig[10]{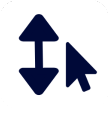} to adjust the contribution score of selected bins by dragging them.
    Similarly, the user can apply the Interpolate tool \inlinefig[10]{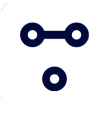} to linearly interpolate the scores of an interval of bins.
  }
  \label{fig:toolbar}
\end{figure}

\textbf{Editing tools.}
Once a region of the shape function is selected, the \toolbar{}~(\autoref{fig:toolbar}) appears: it affords a variety of editing functions, represented as icon buttons, to meet editing needs in different scenarios.
Users can click the buttons to change the shape function in the selected regions.
For example, a user can click the monotonically increasing button \inlinefig{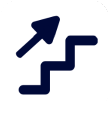} to transform an interval of the shape function into a monotonically increasing function.
In some settings, such monotonicity may be required, for example by regulations for financial ML models.
Internally, \tool{} fits an isotonic regression~\cite{barlowStatisticalInferenceOrder1972} weighted by the bin counts, and uses it to determine a new shape function that meets the regional monotonicity constraint with minimal changes.
The user can view the number of samples in the selected bins and a description of their edit on the bottom \statusbar{}~(\autoref{fig:teaser}A).
Then, they can click the check icon \inlinefig{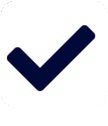} to ``commit''~(\autoref{sec:historyview}) the change if they are satisfied with this edit, or click the cross icon \inlinefig{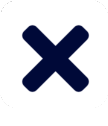} to discard the change.

\toggletrue{inheader}
\subsection{\metricview{}}
\label{sec:metricview}
\togglefalse{inheader}

\setlength{\columnsep}{8pt}%
\setlength{\intextsep}{0pt}%
\begin{wrapfigure}{R}{0.19\textwidth}
  \vspace{0pt}
  \centering
  \includegraphics[width=0.19\textwidth]{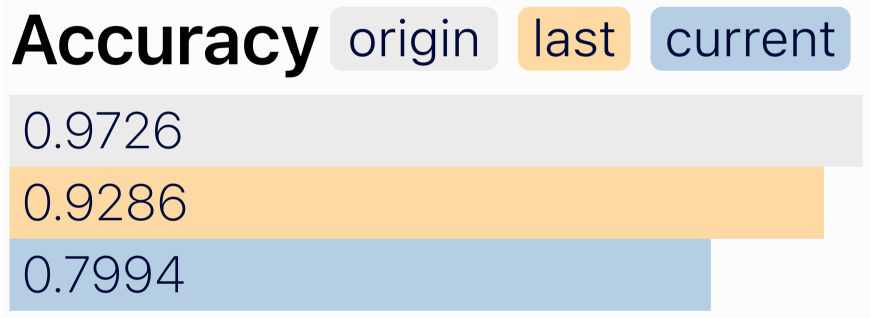}
\end{wrapfigure}

One of our design priorities is to promote responsible editing.
To help users identify the effects of their edits, the \metricview{}~(\autoref{fig:teaser}-B1) provides real-time and continuous feedback on the model performance.
For a binary classifier, we present a confusion matrix and use bar plots to encode the model's accuracy, balanced accuracy, and the Area Under the Curve (AUC).
Similarly, for a regression model, we report root mean squared error, mean absolute error, and mean absolute percentage error.
Also, we use the same color codes of shape functions in the \canvasview{} to describe the performance of the original model, the model from the last edit, and the current model.

\setlength{\columnsep}{8pt}%
\setlength{\intextsep}{0pt}%
\begin{wrapfigure}{R}{0.19\textwidth}
  \vspace{3pt}
  \centering
  \includegraphics[width=0.19\textwidth]{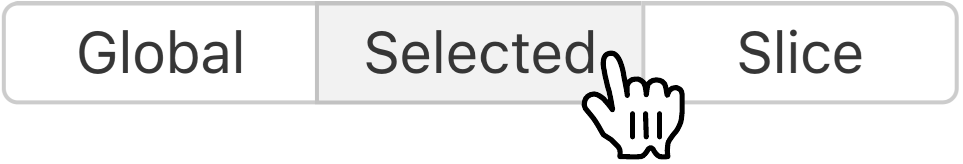}
\end{wrapfigure}

Besides monitoring global metrics that are computed on all test samples, users can change the metric scope by clicking the \scopeswitch{}.
For example, in the \textit{Selected Scope}, the \metricview{} only reports the model performance of samples that are in the selected region.
It helps users to focus on samples that are affected by their edits.
In the \textit{Slice Scope}, users can slice the samples by selecting a level of a categorical variable, e.g., the \inlinefig[9]{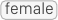} level of the \inlinefig[9]{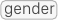} variable.
Then, performance metrics are computed on the test samples that belong to the selected subgroup.
Allowing users to study the model performance on different slices of the dataset provides an opportunity to probe for equitable editing across different groups~\cite{wexlerWhatIfToolInteractive2019}.

\toggletrue{inheader}
\subsection{\featureview{}}
\label{sec:featureview}
\togglefalse{inheader}

\begin{figure}[tb]
  \includegraphics[width=\linewidth]{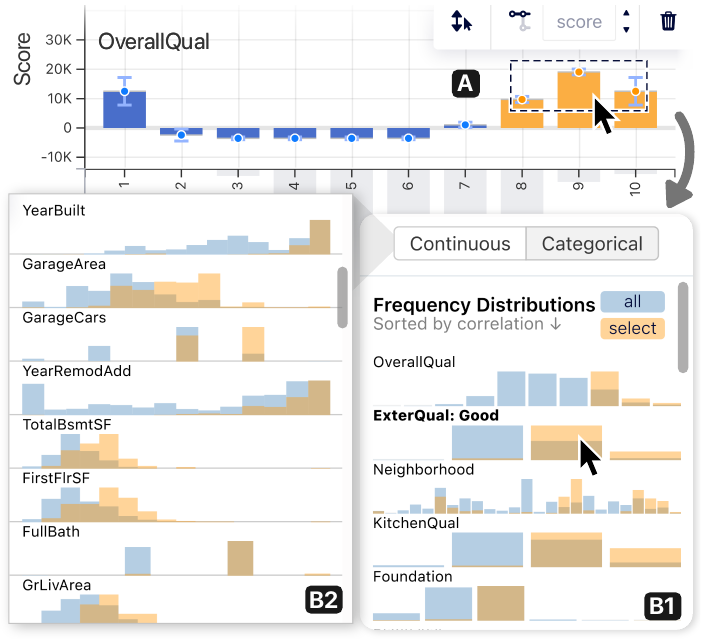}
  \Description{User interface for \tool{}, featuring four tightly integrated views:
    \canvasview{}, \metricview{}, \featureview{}, and \historyview{}
  }
  \caption{
    \inlinefig[10]{a2} On a GAM trained to predict house price, a user selects bins representing high house quality in the \canvasview{}.
    \inlinefig[10]{b2-1} For categorical variables, the \featureview{} shows that selected houses disproportionally have better exterior and kitchen quality and locate in certain neighborhoods.
    \inlinefig[10]{b2-2} For continuous variables, the year built and garage area also highly correlate with the house quality.
  }
  \label{fig:correlation}
\end{figure}

Feature correlations and confounders convolute ML interpretation~\cite{molnarGeneralPitfallsModelAgnostic2021}.
Glass-box models like GAMs can help people identify missing confounders by revealing the effects of multiple features~\cite{rudinInterpretableMachineLearning2021}.
Due to space constraints, in the \canvasview{} users can only inspect and edit one feature at a time.
Therefore, we design the \featureview{}~(\autoref{fig:teaser}-B2, \autoref{fig:correlation}) to help users gain an overview of correlated features as well as their distributions and elucidate potential editing impact disparities.
Inspired by visualization techniques \textit{context+focus}~\cite{cardReadingsInformationVisualization1999} and \textit{brushing+linking}~\cite{bujaInteractiveDataVisualization1991}, we develop \textbf{linking+reordering}---a simple method to identify correlated features.

Once a user selects an interval of the shape function in the \canvasview{}~(\autoref{fig:correlation}-A), we look up affected samples and their associated bins across all features.
For each feature, we compare the bin count frequency of \bluehl{all training data} and the frequency of the \orangehl{\strut selected samples} by measuring the $\ell_2$ distance between these two frequency vectors.
Then, we plot two frequency distributions in an overlaid histogram for each feature, and sort all histograms in descending order of the distance scores~(\autoref{fig:correlation}-B).
The intuition is that if two features $x_1$ and $x_2$ are independent, then samples selected from an interval in $x_1$ should have a \orangehl{distribution} similar to the \bluehl{training data distribution} in $x_2$, and vice versa.
Therefore, correlated features will be on top of the sorted histogram list.
Our \textit{linking+reordering} technique allows users to interactively and quickly identify local correlations across features, even between continuous and categorical features.
By observing correlated features and their distributions, users can identify potential editing effect disparity: for example, editing in \autoref{fig:correlation} would disproportionately affect newer houses.

\toggletrue{inheader}
\subsection{\historyview{}}
\label{sec:historyview}
\togglefalse{inheader}

\setlength{\columnsep}{5pt}%
\setlength{\intextsep}{0pt}%
\begin{wrapfigure}{R}{0.1\textwidth}
  \vspace{0pt}
  \centering
  \includegraphics[width=0.1\textwidth]{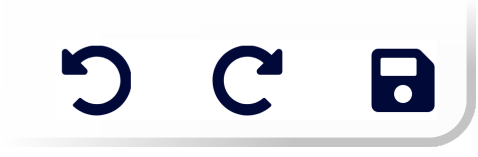}
\end{wrapfigure}

In \tool{}, users can undo and redo their edits by clicking the buttons in the bottom \statusbar{} or using keyboard shortcuts.
Reversible model editing promotes accountable modifications, as users can easily fix their editing mistakes.
To empower users to quickly and easily trace, compare, and revert the changes they have made, the \historyview{} tracks all edits and shows each one as an information card in a list~(\autoref{fig:teaser}-B3).
Inspired by the version control system Git\footnote{Git: \link{https://git-scm.com}}, we model each edit as a commit---a snapshot of the underlying GAM.
Each commit has a timestamp, a unique identifier, and a commit message.

\setlength{\columnsep}{9pt}%
\setlength{\intextsep}{0pt}%
\begin{wrapfigure}{R}{0.2\textwidth}
  \vspace{0pt}
  \centering
  \includegraphics[width=0.2\textwidth]{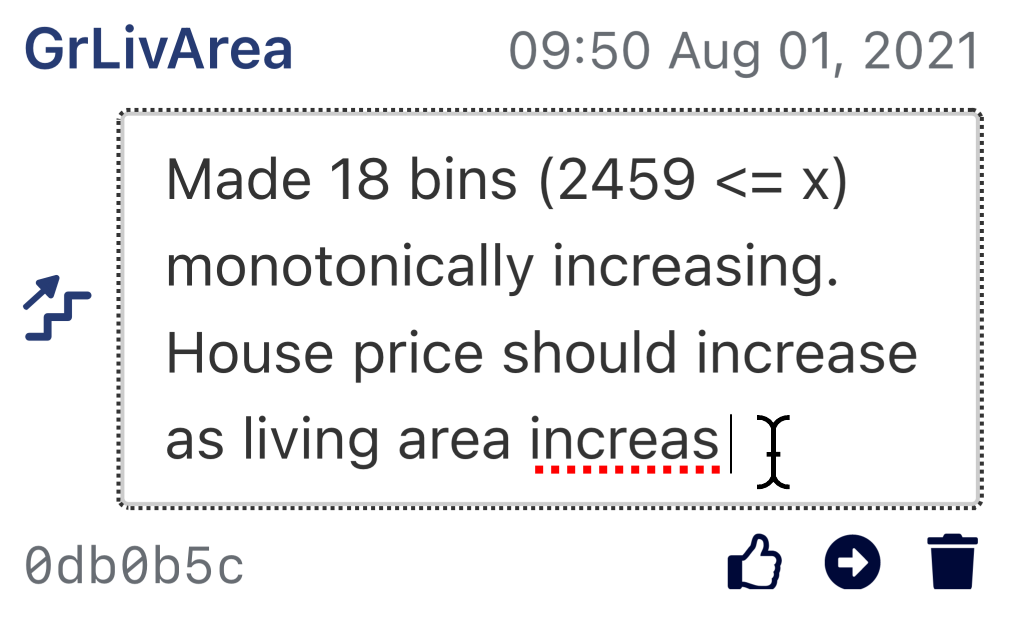}
\end{wrapfigure}

Once an edit is committed, we automatically generate an initial commit message to describe the edit; users can update the message in the \historyview{} to further document their editing motivation and context.
In addition, users can browse the model evolution by clicking the Check-out button \inlinefig{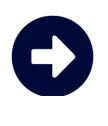} to load previous GAM versions.
Users can also discard edits by clicking the Delete button \inlinefig{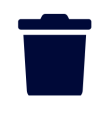}.
Once users finishes editing, they can click the Save button \inlinefig{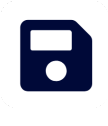} in the \statusbar{} to save the latest GAM along with all editing history, which can be used for deployment or future continuing editing.
Before saving the model, \tool{} requires users to examine and confirm all edits by clicking the Confirm button \inlinefig{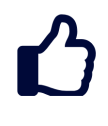}.
It helps users identify editing mistakes and promote accountable editing~\cite{gerlachUnderstandingHumanComputerInteraction1991}.

\toggletrue{inheader}
\subsection{Web-based, Open-source Implementation}
\label{sec:implementation}
\togglefalse{inheader}

\tool{} is a modern web-based interface built with \textit{D3.js}~\cite{bostockDataDrivenDocuments2011}.
It is compatible with the popular GAM library \textit{InterpretML}~\cite{noriInterpretMLUnifiedFramework2019}: users can easily export GAMs to edit and load modified GAMs.
We use WebAssembly to accelerate in-browser computations, such as GAM inference and isotonic regression.
It makes our tool scalable: all computations are real-time with less than 5k test samples, and the sample size is only bounded by the browser memory limit.
Users can also use our tool directly in their computational notebooks---popular workflow for ML development~\cite{zhangHowDataScience2020}.
With modern web technologies, our tool makes it accessible for domain experts and data scientists to locally and privately edit GAMs with minimal coding.
We open source \tool{} and all WebAssembly components, so future researchers and designers can easily generalize our design and implementations to other forms of model editing. %
\section{Usage Scenario}
\label{sec:scenario}

We present a hypothetical usage scenario to illustrate how \tool{} could potentially help doctors and data scientists discover and fix harmful behaviors in a GAM that predicts a patient’s risk of dying from pneumonia.
In this usage scenario, all metrics reported are based on models trained on a dataset of 14,199 pneumonia patients~\cite{caruanaIntelligibleModelsHealthCare2015}.
The dataset has 46 features: 19 features are continuous, while 27 are categorical.
The outcome variable has a binary value: \texttt{1} if the patient died of pneumonia and \texttt{0} if they survived.

\textbf{Model training.}
On the pneumonia dataset, the popular ML models XGBoost~\cite{chenXGBoostScalableTree2016}, Random Forest~\cite{breimanRandomForests2001}, and GAM trained with boosted-trees~\cite{louAccurateIntelligibleModels2013} all have similar performance: Area Under the Curve (AUC) scores of \texttt{0.850}, \texttt{0.845}, and \texttt{0.853}, respectively.
Since in practice, doctors tend to prefer interpretable models like GAMs over black-box models~\cite{wiensNoHarmRoadmap2019}, we expect that doctors would decide to proceed with the GAM for the pneumonia risk prediction task.

\begin{figure}[tb]
  \includegraphics[width=\linewidth]{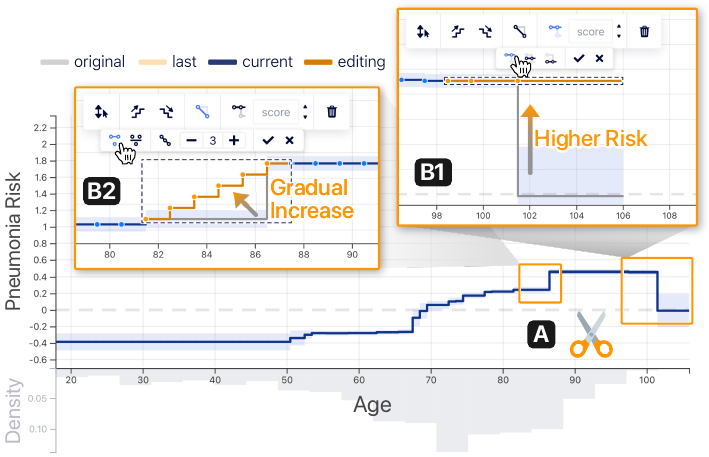}
  \Description{User interface for \tool{}, featuring four tightly integrated views:
    \canvasview{}, \metricview{}, \featureview{}, and \historyview{}
  }
  \caption{
    \inlinefig[10]{a2} Contrary to doctors' knowledge, a GAM predicts patients above \texttt{100} years old to have lower  risk of dying from pneumonia than patients \texttt{20} years younger (right), and there is a rapid increase of risk from age \texttt{86} to \texttt{87} (left).
    \inlinefig[10]{b2-1} By clicking the Align button \inlinefig[10]{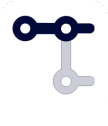}, a doctor can raise the risk score for older patients;
    \inlinefig[10]{b2-2} The doctor can also apply the Interpolation tool \inlinefig[10]{interpolate} to smooth out the abrupt increase of risk.
  }
  \label{fig:scenario1}
\end{figure}

\textbf{Model interpretation.}
Users---in this case, doctors and data scientists---can easily set up \tool{} in a web-based interface with drag-and-drop or directly in their computational notebooks.
\inlinefig[9]{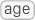} is the first feature \tool{} presents to users, as it has the highest importance score across all features.
In the \canvasview{}~(\autoref{fig:scenario1}A), the x-axis ranges from \texttt{18} to \texttt{106} years old.
The y-axis encodes the predicted risk score (log odds) of dying from pneumonia.
It ranges from a score of \texttt{-0.4} for patients in their 20s to \texttt{0.5} for patients in their 90s.
A negative risk score represents lower risk than the average patients, while a positive score means higher risk than the average patients.
The light blue background below the line plot encodes the model's uncertainty, which is strongly correlated with the training sample size in different \inlinefig[9]{tag-age} ranges, as shown in the histogram at the bottom of the \canvasview{}.

By observing the shape function visualization~(\autoref{fig:scenario1}A), users can see that the model predicts younger patients to have a lower risk than elder patients.
The predicted risk rapidly increases when patients get older than \texttt{67}.
However, the predicted risk suddenly plunges when the \inlinefig[9]{tag-age} passes \texttt{100}---leading to a similar risk score as if the patient is \texttt{30} years younger!
This prediction pattern also exists in the XGBoost and Random Forest models, but is harder to detect since these models are black-box.

There are several hypotheses to explain this dangerous behavior across ML models.
For example, it might be due to outliers in this \inlinefig[9]{tag-age} range, especially the range has a small sample size (shown in the histogram), or patients who live this long might have ``good genes'' to recover from pneumonia.
To identify the true impact of \inlinefig[9]{tag-age} on pneumonia risk, additional causal experiments and analysis are needed.
Without robust evidence that people over 100 are truly at lower risk, many doctors may fear that they would be injuring patients by depriving needy older people of care, and violating their primary obligation to \textit{do no harm}.
Therefore, doctors would like to fix this pattern.
A conservative remedy is to set the risk of older patients to be equal to that of their slightly younger neighbors.

\setlength{\columnsep}{6pt}%
\setlength{\intextsep}{0pt}%
\begin{wrapfigure}{R}{0.2\textwidth}
  \vspace{0pt}
  \centering
  \includegraphics[width=0.2\textwidth]{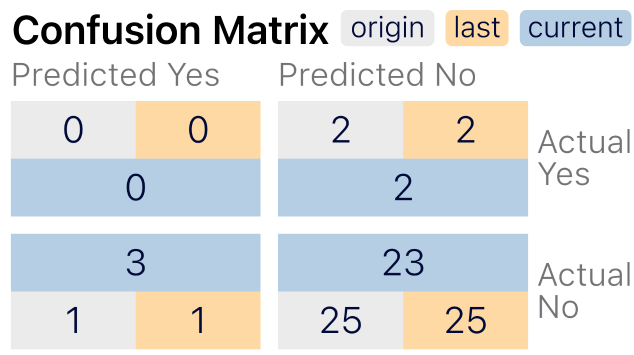}
\end{wrapfigure}

\textbf{Editing continuous features.}
To do that, a doctor can select the bins that represent \inlinefig[9]{tag-age} above 98 years old through drawing a bounding box in the \canvasview{}~(\autoref{fig:scenario1}-A).
Then, the \toolbar{} appears, where the doctor can click the Align to The Left button\inlinefig{align}~(\autoref{fig:scenario1}-B1).
With a smooth animation, \tool{} raises the bin that encodes \inlinefig[9]{tag-age} above \texttt{100} vertically, so that it has the same risk score as the first bin on the left (\inlinefig[9]{tag-age} \texttt{98-100}).
From the \metricview{}, the doctor can see that the accuracy decreases by \texttt{0.0004} when evaluating all 5000 test samples in the \textit{Global Scope}~(\autoref{sec:metricview}).
To focus on the model performance on the affected 28 samples, the doctor can switch to the \textit{Selected Scope}: the confusion matrix shows that the edit causes the model to mis-classify two negative cases as false positives.

\setlength{\columnsep}{6pt}%
\setlength{\intextsep}{0pt}%
\begin{wrapfigure}{R}{0.2\textwidth}
  \vspace{0pt}
  \centering
  \includegraphics[width=0.2\textwidth]{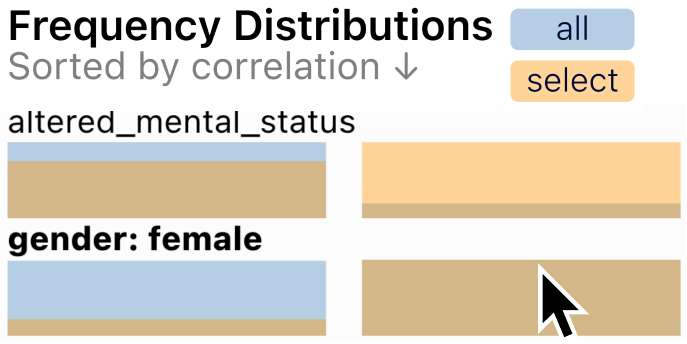}
\end{wrapfigure}

The accuracy drop is negligible, as any change to an optimized model is likely to hurt the performance on the training data; the edit is likely to make the model generalize better to unseen data.
To learn more about these 28 patients who would be affected by the edit, the doctor can open the \featureview{}, which shows that \inlinefig[9]{tag-gender} is the second most correlated categorical feature with the selected \inlinefig[9]{tag-age} range.
It means patients who are affected by this edit are disproportionally female---it makes sense because on average women live longer than men.
By identifying and showing the correlated features, the \featureview{} promotes users to be aware of potential fairness issues during model editing.

Besides the problematic drop of predicted risk for older patients, the risk suddenly rises around \inlinefig[9]{tag-age} \texttt{86}~(\autoref{fig:scenario1}A).
After converting the risk score from the log-odds space to probability space, the predicted likelihood of dying from pneumonia increases by $4.89\%$ when the \inlinefig[9]{tag-age} goes from \texttt{86} to \texttt{87}!
Similar to the previous discussion, this model behavior can cause \texttt{81}--\texttt{86} years old patients miss the care they need.
To fix this pattern, the doctor can select the region from \inlinefig[9]{tag-age} \texttt{81} to \texttt{87} and click the Interpolate button\inlinefig{interpolate}~(\autoref{fig:scenario1}-B2).
Then, \tool{} would linearly interpolate risk scores in the selected range, so that the model behavior matches doctors' domain knowledge of a gradual risk increase from \inlinefig[9]{tag-age} \texttt{81} to \texttt{87}.

\begin{figure}[b]
  \includegraphics[width=\linewidth]{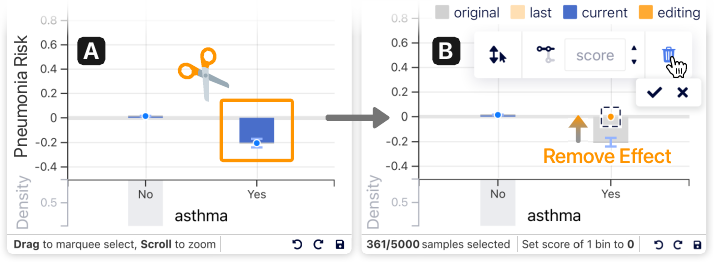}
  \Description{User interface for \tool{}, featuring four tightly integrated views:
    \canvasview{}, \metricview{}, \featureview{}, and \historyview{}
  }
  \caption{
    \inlinefig[10]{a2} As opposed to doctors' knowledge, a GAM predicts having asthma lowers the risk of dying from pneumonia.
    \inlinefig[10]{b2} A doctor can repair this harmful pattern by applying the Delete tool \inlinefig[10]{delete} to remove the effect of having asthma.
  }
  \label{fig:scenario2}
\end{figure}

\textbf{Editing categorical features.}
In addition to continuous features, \tool{} also allows users to edit categorical features.
For example, the \canvasview{} of the binary feature \inlinefig[9]{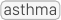} shows that the model predicts asthmatic patients to have lower pneumonia risk than non-asthmatic patients~(\autoref{fig:scenario2}A).
This model behavior is likely to go against doctors' knowledge and experience.
A conservative option to modify this behavior is to remove the predictive effect of having \inlinefig[9]{tag-asthma}, so that the model predicts asthmatic patients to have an average risk.
To do that, a doctor can select the bar in the \canvasview{} and clicks the Delete button \inlinefig{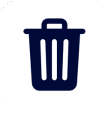}.
It would increase the risk score of having \inlinefig[9]{tag-asthma} from \texttt{-0.2} to \texttt{0}.
Then, the doctor can document the edit motivation and context for future references by leaving a comment, such as ``Conservative way to fix the wrong risk score of having \inlinefig[9]{tag-asthma},'' in the \historyview{}.
Finally, after reviewing and confirming all edits by clicking the Confirm button \inlinefig{confirm}, the doctor can save the new model and edit history.

\section{Ongoing Work and Conclusion}

With a responsible
and familiar
design, \tool{} enables domain experts and data scientists to turn ML interpretability into actions.
From the usage scenario~(\autoref{sec:scenario}), we can see that model editing is an iterative and complex process, and it is new to many ML researchers and practitioners.
To help \tool{} users create more accountable ML models, our ongoing work includes:

\begin{itemize}[topsep=1mm, itemsep=0mm, parsep=1mm, leftmargin=3mm]
  \item \textbf{Evaluate the usability} of \tool{} by running an observational user study with data scientists.
  \item \textbf{Case studies} on real datasets with doctors and data scientists.
  We will deploy edited GAMs in healthcare settings and monitor their performance in the real life.
  \item \textbf{Distill guiding principles} of \textit{when} we should edit ML models and \textit{how} to edit them in different scenarios.
\end{itemize}

\noindent \textbf{Limitations.}
We design \tool{} to guard against harmful edits by providing users with continuous feedback~(\cref{sec:featureview,sec:metricview}), as well as transparent and reversible edits~(\cref{sec:historyview}).
However, it does not guarantee to prevent users from overfitting the model, injecting harmful bias, or maliciously manipulating model predictions.
This potential vulnerability warrants further study on how to audit and regulate model editing.
We will deploy \tool{} with caution, monitor its impact, and continuously improve the design to promote responsible editing.

\noindent \textbf{Conclusion.}
\tool{} takes steps toward democratizing responsible ML.
Through applying interactive visualization techniques, \tool{} provides an easy-to-use interface that empowers domain experts and data scientists to not only interpret ML models, but also align model behaviors with their knowledge and values.
We hope our work helps emphasize the importance of human agency in responsible ML research, inspiring future work in human-AI interaction and actionable ML interpretability. %
\section{Acknowledgement}

The first two authors were summer interns at Microsoft Research.
We especially thank Scott Lundberg for insightful conversations.
We are also grateful to Steven Drucker, Adam Fourney, Saleema Amershi, Dean Carignan, Rob DeLine, and the InterpretML team for their helpful feedback.
\bibliographystyle{ACM-Reference-Format}
\balance
\bibliography{gam-changer}

\end{document}